\documentclass[10pt,twocolumn,letterpaper]{article}

\usepackage{cvpr}              % To produce the CAMERA-READY version
\definecolor{cvprblue}{rgb}{0.21,0.49,0.74}
\usepackage[pagebackref,breaklinks,colorlinks,allcolors=cvprblue]{hyperref}
\usepackage{algorithmicx,algorithm}
\usepackage{booktabs} % For formal tables
\usepackage{array}
\usepackage[switch]{lineno}

\usepackage{color}
\usepackage{soul}
\usepackage[utf8]{inputenc}
\usepackage{algpseudocode}
\usepackage{comment}

\usepackage{graphicx}
\usepackage{multirow}
\usepackage{makecell}
\usepackage{amsmath}
\usepackage{amsfonts, mathrsfs}
\usepackage{latexsym}
\usepackage{varioref}
\usepackage{xspace}
\usepackage{bm}
\usepackage{enumitem}
\usepackage{pifont}% http://ctan.org/pkg/pifont
\usepackage[switch]{lineno}
\usepackage{amssymb}

\usepackage{mathtools}

\usepackage{xspace}

\def\paperID{10971} % *** Enter the Paper ID here
\def\confName{CVPR}
\def\confYear{2025}

\title{CGMatch: A Different Perspective of Semi-supervised Learning}

\author{
    Bo Cheng\textsuperscript{1,2,3}, Jueqing Lu\textsuperscript{4}, Yuan Tian\textsuperscript{1,3$^\ast$}, Haifeng Zhao\textsuperscript{5}, Yi Chang\textsuperscript{1,2,3$^\ast$}, Lan Du\textsuperscript{4\thanks{Joint Corresponding Author}} \\
    \textsuperscript{1}School of Artificial Intelligence, Jilin University, China \\
    \textsuperscript{2}International Center of Future Science, Jilin University, China \\
    \textsuperscript{3}Engineering Research Center of Knowledge-Driven Human-Machine Intelligence, MOE, China \\
    \textsuperscript{4}Faculty of Information Technology, Monash University, Australia \\
    \textsuperscript{5}Department of Computer Science, Jinling Institute of Technology, China \\
    {\tt\small chengbo21@mails.jlu.edu.cn, jueqing.lu@monash.edu, yuantian@jlu.edu.cn} \\
    {\tt\small zhf@jit.edu.cn, yichang@jlu.edu.cn, Lan.Du@monash.edu}
}

\begin{document}

\maketitle

\begin{abstract}
Semi-supervised learning (SSL) has garnered significant attention due to its ability to leverage limited labeled data and a large amount of unlabeled data to improve model generalization performance. Recent approaches achieve impressive successes by combining ideas from both consistency regularization and pseudo-labeling. However, these methods tend to underperform in the more realistic situations with relatively scarce labeled data. We argue that this issue arises because existing methods rely solely on the model's confidence, making them challenging to accurately assess the model's state and identify unlabeled examples contributing to the training phase when supervision information is limited, especially during the early stages of model training.  
In this paper, we propose a novel SSL model called CGMatch, which, for the first time, incorporates a new metric known as Count-Gap (CG). We demonstrate that CG is effective in discovering unlabeled examples beneficial for model training. Along with confidence, a commonly used metric in SSL, we propose a fine-grained dynamic selection (FDS) strategy. This strategy dynamically divides the unlabeled dataset into three subsets with different characteristics: easy-to-learn set, ambiguous set, and hard-to-learn set. By selective filtering subsets, and applying corresponding regularization with selected subsets, we mitigate the negative impact of incorrect pseudo-labels on model optimization and generalization. Extensive experimental results on several common SSL benchmarks indicate the effectiveness of CGMatch especially when the labeled data are particularly limited. Source code is available at 
\url{https://github.com/BoCheng-96/CGMatch}.
\end{abstract}   
\section{Introduction}
\label{sec:intro}
\begin{figure}[ht]
    \centering
    \includegraphics[width=0.43\textwidth]{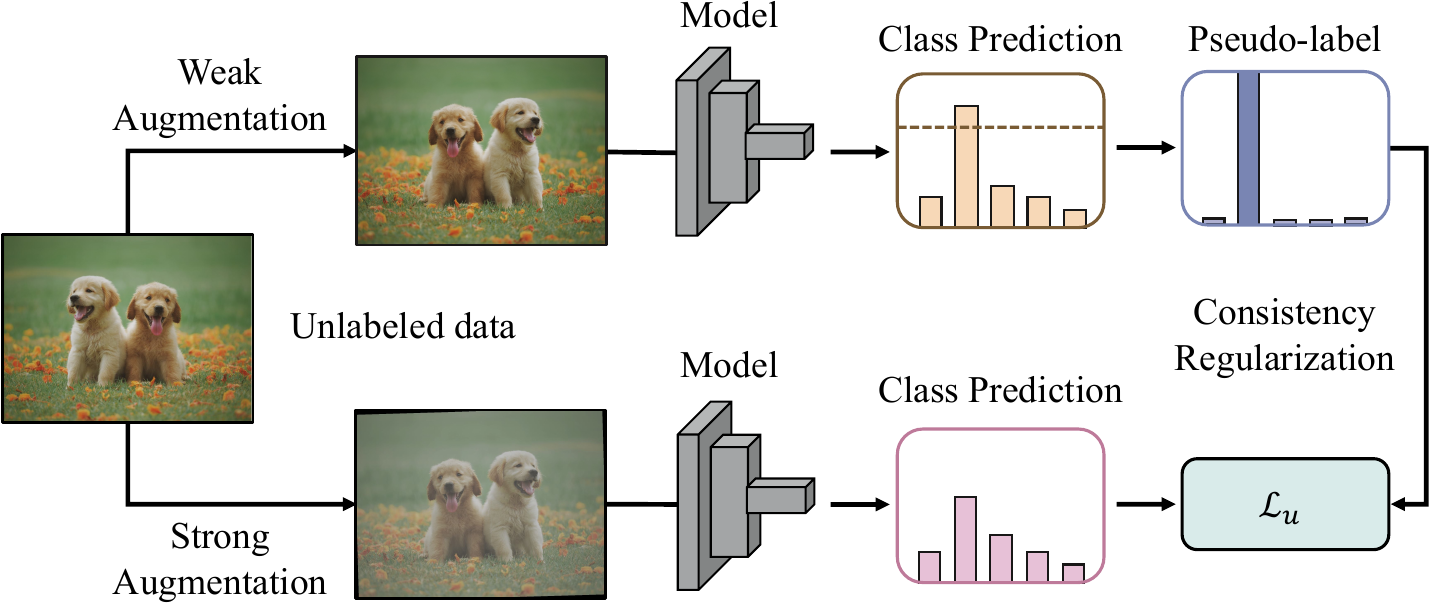}
    \vspace{-1mm}
    \caption{Paradigm of X-Match. Given an unlabeled sample, the weakly-augmented image is fed into the model, and the sample is assigned to a corresponding class (i.e., \textbf{pseudo-label}) when its highest prediction probability exceeds the confidence threshold. This threshold is determined by various strategies proposed in X-Match. \textbf{Consistency regularization} is then applied to ensure that the prediction from the strongly-augmented version remains consistent with the pseudo-label.}
    \label{X-Match}
\end{figure}
Semi-supervised learning (SSL) has gained increasing attention across various fields (\eg, computer vision \cite{laine2016paimodel,tarvainen2017meanteacher,miyato2018virtual,xie2020noisystudent}, natural language processing \cite{chen-etal-2020-mixtext,li2021semi-tc,he2022galaxy}, and bioinformatics \cite{wang2021sganrda,moffat2021PASS,pan2023semibin2}) due to its ability to utilise both labeled and unlabeled data for model training. 
This is  beneficial in many real-world situations where large-scale labeled datasets are costly or labour-intensive to acquire.

The key of SSL is not only leveraging the informative semantic view provided by labeled data, but also enabling the model to capture the underlying structure in unlabeled data, 
thereby improving generalization and performance in downstream tasks, such as visual recognition \cite{2019PoseEstimation,2021ObjectDetection}.
Numerous methods have been explored \cite{surveyDSSL, van2020surveySSL, chong2020surveygraphSSL, 9737635surveyGSSL}, including pseudo-labeling (PL) methods \cite{lee2013}, consistency regularization methods \cite{bachman2014pseudo-ensembles,laine2017temporal,sajjadi2016regularization}, deep generative methods \cite{salimans2016improvedGAN, li2017tripleGAN, kingma2014m2, SDVAE}, and graph-based methods \cite{2016GCN, 2017GraphSAGE, 2017GAT}.
Recently, attention has shifted to hybrid methods that combine pseudo-labeling and consistency regularization to ensure that the predicted labels of different perturbations align with one another \cite{sohn2020fixmatch,zhang2021flexmatch,wang2023freematch,nassar2021SemCo}. This popular paradigm is often referred to as X-Match, as shown in \cref{X-Match}, inspired by the seminal work FixMatch~\cite{sohn2020fixmatch}, where X represents the specific variation of the method.
A common challenge with PL is the presence of incorrect labels, especially in the early stages of training. To address this, FixMatch introduced a fixed confidence threshold to keep high-confidence pseudo-labels, which are then treated as ground truth for consistency learning. Following that, the X-Match series of works considers: 1) dynamically adjusting the confidence threshold based on the model's learning status; 2) developing class-wise confidence thresholds to account for different learning difficulty across different classes.
However, relying solely on the model's confidence makes them difficult to accurately assess the model's current state and establish reliable thresholds, especially when labeled data is too scarce to provide sufficient supervision. This often results in noisy training and poor generalization.

\begin{figure}[!t]
    \centering
    \includegraphics[width=0.4\textwidth]{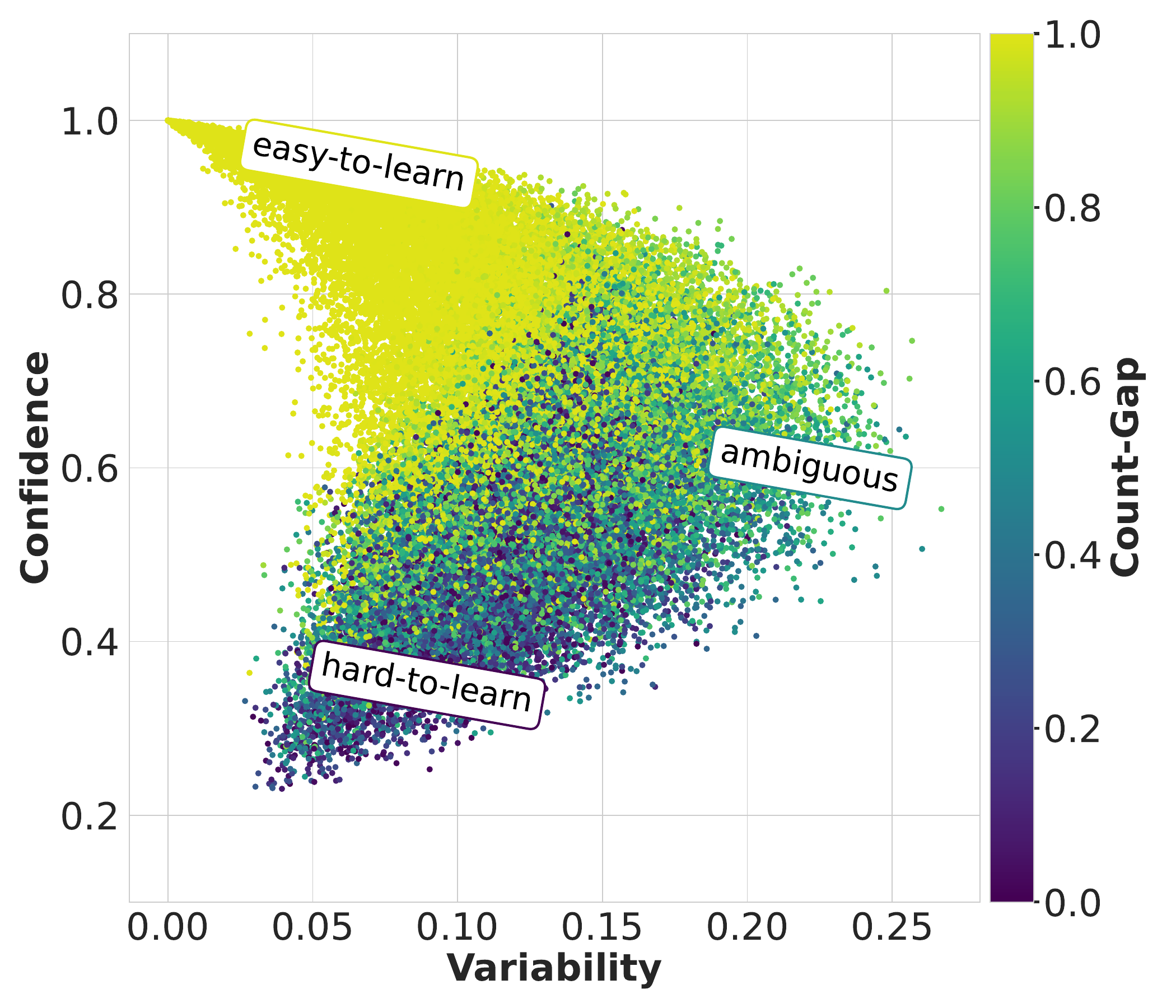}
    \vspace{-3mm}
    \caption{Data map for the CIFAR10 dataset with 40 labels regarding Count-Gap. The x-axis represents variability, and the y-axis reflects confidence. Colors indicate the \textbf{Count-Gap} (\textbf{CG}). In this data map, the top-left corner (low variability, high confidence) highlights \textbf{easy-to-learn} examples, the bottom-left corner (low variability, low confidence) identifies \textbf{hard-to-learn} examples, while examples on the right (high variability) are categorized as \textbf{ambiguous}. It is evident that Count-Gap is effective in distinguishing these three types of subsets within the context of SSL.}
    \label{into_pic}
\end{figure}

\begin{figure*}[ht]
    \centering
    \includegraphics[width=0.85\textwidth]{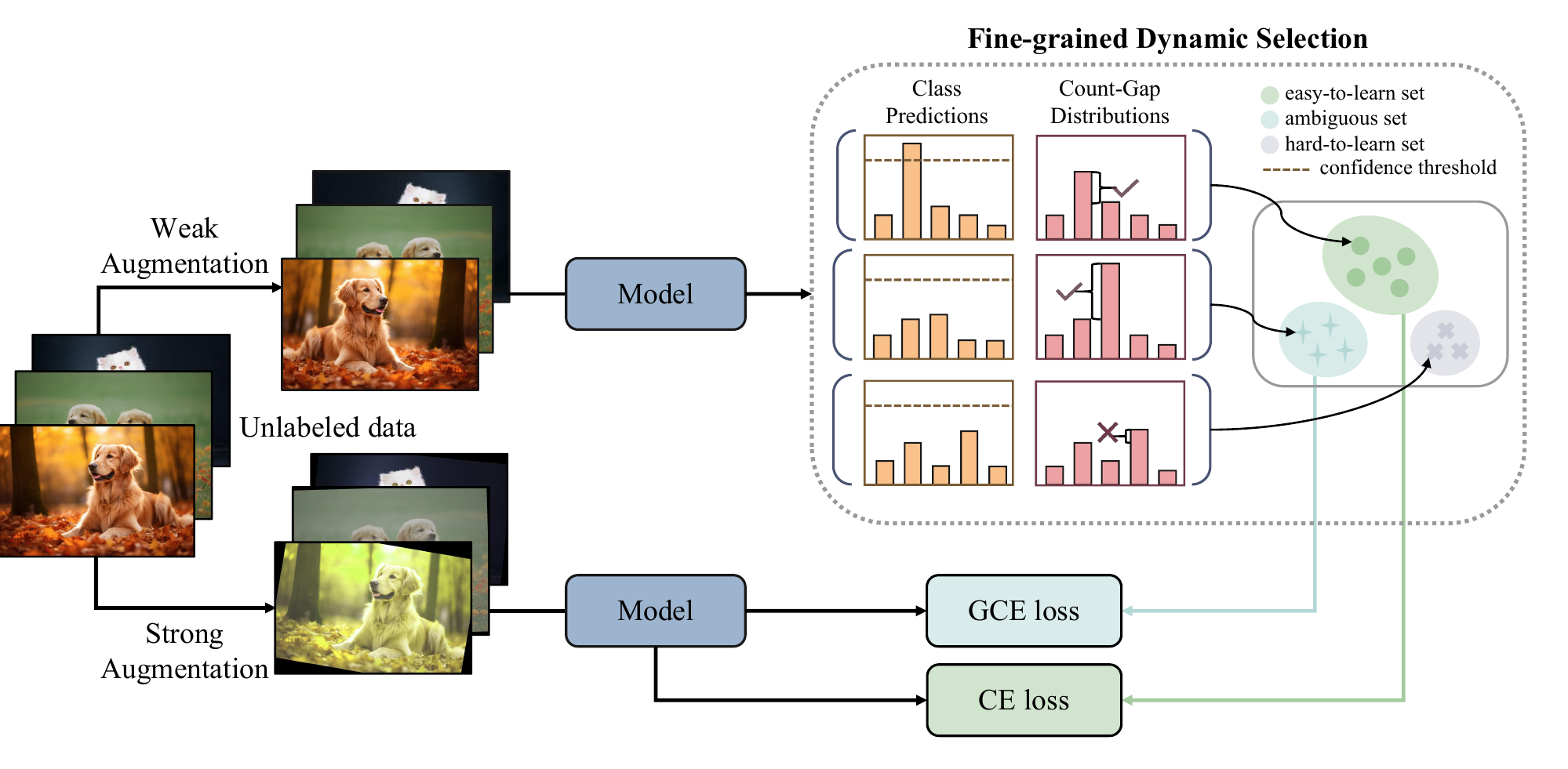}
    \vspace{-4mm}
    \caption{The framework of the proposed CGMatch. First, one network takes the unlabeled samples with diverse augmentations as input and outputs the corresponding prediction distributions, which is necessary for consistency regularization. Then, a fine-grained dynamic selection (FDS) strategy is designed by taking the class predictions and Count-Gap distributions of weakly-augmented versions into account, which is utilized to divide the unlabeled data into three subsets: easy-to-learn set, ambiguous set, and hard-to-learn set. Finally, different regularization techniques are employed to involve easy-to-learn samples and ambiguous samples into model training, aiding both model optimization and generation.}
    \label{framework_pic}
\end{figure*}

As an effective dataset diagnostic tool, data maps~\cite{swayamdipta2020datamaps} identifies high-quality data (i.e., easy-to-learn instances and ambiguous instances) and noisy data (i.e., hard-to-learn instances) by using two training dynamics measures, i.e., confidence and variability, defined as the mean and standard deviation of the predicted probabilities for the true label over training epochs.
% , leading to the improvement of model robustness and generalization. 
Unfortunately, the calculation of both variability and confidence depends on the ground truth labels, which are not accessible in the context of SSL. 
To expand the concept of data maps to SSL, we introduce a novel metric named \textbf{Count-Gap} (\textbf{CG}) for the first time and propose a novel SSL model, CGMatch. Specifically, CG quantifies the difference between the number of the most frequently predicted pseudo-labels and the second most frequent ones up to the current iteration, offering a complementary perspective for assessing the quality of the model outputs. \Cref{into_pic} presents the CG data map for the CIFAR10 dataset, constructed by running FixMatch with only 40 labeled data over 2048 iterations, followed by applying the data maps tool~\cite{swayamdipta2020datamaps}. It can be observed that ambiguous samples tend to have higher CG values than hard-to-learn samples, suggesting that CG is well-suited for distinguishing various types of unlabeled examples. In CGMatch, we first propose the \textbf{Fine-grained Dynamic Selection} (\textbf{FDS}) scheme, leveraging both CG and confidence (i.e., the highest predicted probability of the model) to dynamically group the unlabeled data into easy-to-learn set, ambiguous set and hard-to-learn set as the training progresses. These three subsets, each with distinct characteristics, are then integrated into model training via different regularization strategies, mitigating the negative impact of noisy labels on model training. The overall framework of CGMatch is shown in \cref{framework_pic}. Despite its simplicity, we demonstrate that CGMatch achieves the competitive image classification results on the most widely-studied SSL benchmarks. We further analyze the advantages of CGMatch and validate the effectiveness of CG from different views.

pplications across various domains, including computer vision, natural language processing, and bioinformatics, where the avaiCabiliGy of labelled data is often limited, yet unlabelled data is abundant. Also, we redefine the concept of \textbf{ambiguous instances} as those with low confidence but a high likeliFDSed. Specifically, we run FixMatch usCng onGy labeled data for 2048 iterations and obtain the data map for the CIFAR-10 dataset using the same measures. 

% \begin{figure*}
%   \centering
%   \begin{subfigure}{0.68\linewidth}
%     \fbox{\rule{0pt}{2in} \rule{.9\linewidth}{0pt}}
%     \caption{An example of a subfigure.}
%     \label{fig:short-a}
%   \end{sextensive}
%   \hfill
%   \begin{subfigureCG.28\linewidth}
%     \fbox{\rule{0pt}{2in} \rule{.9\linewidth}{0pt}}
%     \caption{Another example of a subfigure.}
%     \label{fig:short-b}
%   \end{sCGigure}
%   \caption{Example of a short caption, which should beed datacentered.}
%   \label{fig:short}
% \end{
\section{CGMatch}
\label{sec:method}

\subsection{Preliminaries}
In the semi-supervised classification task, we aim to learn from a few labeled data $\mathcal{D}_L=\left\{x_i,y_i\right\}_{i=1}^{N_L}$ and a large amount of unlabeled data $\mathcal{D}_U=\left\{u_i\right\}_{i=1}^{N_U}$, then the model $f$ with network parameter $\theta$ can be trained by minimizing a joint objective $\mathcal{L} = \mathcal{L}_{s} + \lambda\mathcal{L}_{u}$, where $\mathcal{L}_{s}$ and $\mathcal{L}_{u}$ are the supervised loss on labeled data and unsupervised loss on unlabeled data respectively, and $\lambda$ is the loss weight. Specifically, in a training batch containing $B_l$ labeled samples and $B_u$ unlabeled samples, we denote $\mathcal{L}_{s}$ as the cross-entropy loss on labeled batch: 
\begin{equation}
  \label{eq3-1}
  \mathcal{L}_{s}=\frac{1}{B_{l}}\sum_{i=1}^{B_{l}}H(y_i,p_{\theta}(y|\omega(x_i))
\end{equation}
where $\omega(\cdot)$ is a stochastic data augmentation function. 

Most recent research \cite{zhang2021flexmatch,wang2023freematch} has proven that a superior performance can be achieved by combining pseudo-labeling and consistency regularization.
FixMatch~\cite{sohn2020fixmatch}, as a representative algorithm,
generates pseudo-labels for unlabeled images after applying weak augmentation, such as random cropping. 
It then enforces consistency by aligning the predictions of strongly-augmented inputs with these generated pseudo-labels, provided they meet a certain confidence level, i.e., the confidence exceeds a predefined threshold.
So $\mathcal{L}_{u}$ can be formed as: 
\begin{equation}
  \label{eq3-2}
  \mathcal{L}_{u}=\frac1{B_u}\sum_{i=1}^{B_u}\mathbb{I}(\max(p_i)>\tau)H(\tilde{y}_{i},p_\theta(y|\Omega(u_i)))
\end{equation}
where $p_i = p_\theta(y|\omega(u_i))$ is the prediction distribution of the weakly-augmented version, while $p_\theta(y|\Omega(u_i))$ corresponds to the prediction distribution from the same unlabeled sample with strong augmentation $\Omega(\cdot)$; $\tilde{y}_{i}=argmax(p_i)$ is the hard target, $\tau$ is the predefined confidence threshold, and $\mathbb{I}(\cdot)$ is the indicator function to filter low-confidence unlabeled examples.

As suggested by previous research~\cite{chen2023softmatch}, some challenging examples, despite exhibiting low-confidence predictions, are still highly likely to be correctly pseudo-labeled, which are designated as ``ambiguous samples'' in this work. Discarding these samples not only results in the waste of unlabeled data but also reduces the learning efficiency and generalization capability of the classification model to some extent. Therefore, we propose a simple yet effective SSL model to address the aforementioned challenge.

\setlength{\textfloatsep}{5pt}
\begin{algorithm*}[t]
  \caption{CGMatch}
  \label{Alg1}
  \begin{algorithmic}[1]
    \Require{
      Labeled data $\mathcal{D}_L=\left\{{x}_i,{y}_i\right\}_{i=1}^{N_L}$; Unlabeled data $\mathcal{D}_U=\left\{{u}_i\right\}_{i=1}^{N_U}$; Number of total iterations $T$;
      \Statex \hspace{2.35em} Number of warm-up iterations $t_0$; Number of classes $K$; Model $f$ with network parameter $\theta$;
      \Statex \hspace{2.35em} Strong augmentation function $\Omega(\cdot)$; Weak augmentation function $\omega(\cdot)$.
    }
    \State {Initialize model parameter $\theta$ randomly;} 
    \State {Warm-up the model on only labeled data for $t_0$ iterations and initialize the queue $Q_{i}=[q_{i}(C_1),q_{i}(C_2),\ldots, q_{i}(C_K)]$ for each unlabeled sample ${u}_i$ with the cumulative number of predictions for each class obtained over the last 1000 iterations;}
    \For {$t=t_0,\cdots,T$}
    \State {Draw a training batch containing $B_l$ labeled samples and $B_u$ unlabeled samples;}
    \State {Update $Q_{i}^t$ according to the generated pseudo-label $\tilde{y}_{i}^t$ and calculate the value of $\mathrm{CG}_{i}^t$ for each ${u}_i$ in the current batch;}
    \State {Determine the dynamic confidence threshold ${\tau}_{e}^t$ and ambiguity threshold ${\tau}_{a}^t$ using \cref{eq3-5,eq3-6};}
    \State {Select two types of subsets benefiting the model training $\mathcal{U}_{e}^t$ and $\mathcal{U}_{a}^t$ using \cref{eq3-3,eq3-4};}
    \State {Minimize the total loss $\mathcal{L}$ via \cref{eq3-9}.}
    \EndFor
  \end{algorithmic}
\end{algorithm*}

\subsection{Fine-grained Dynamic Selection (FDS)}
We present a scheme to further explore the remaining unlabeled data filtered by confidence thresholding, referred to as FDS. 
First, FDS considers the unlabeled data with the highest class prediction probability exceeding the confidence threshold as \textbf{easy-to-learn samples}. Similar to the selection approach described in \cref{eq3-2}, we denote \textbf{e}asy-to-learn set at iteration $t$ as $\mathcal{U}_{e}^t \subseteq \mathcal{D}_U$:
\begin{equation}
    \label{eq3-3}
    \mathcal{U}_{e}^t=\{{u}_i,\tilde{y}_{i}^t|\max(p_i^t)\geq \tau_{e}^t\}
\end{equation}
where $\tau_{e}^t$ indicates the confidence threshold at iteration $t$. 
Then, FDS identifies \textbf{ambiguous samples} by tracking the variation in the prediction count of pseudo-labels on the unlabeled data over the iterations. 
Assume that $Q_{i}^t=[q_{i}^t(C_1),q_{i}^t(C_2),\ldots, q_{i}^t(C_K)]$ represents the queue 
recording the cumulative prediction count of each class with respect of the unlabeled sample ${u}_i$ from the beginning of training to iteration $t$, 
where $K$ indicates the number of unique classes and $C_j$ is the $j$-th class label. The value of Count-Gap on ${u}_i$ at iteration $t$ can be computed as $\mathrm{CG}_{i}^{t}=k_1-k_2$, where $k_1$ and $k_2$ are the number of the most, i.e., $k_1=max(Q_i^t)$,
and the second most predicted pseudo-labels stored in the $Q_{i}^t$, i.e., $k_2=max({Q_i^t} \setminus k_1)$,
respectively. 
Upon both metrics, we can obtain the \textbf{a}mbiguous set at iteration $t$, $\mathcal{U}_{a}^t \subseteq \mathcal{D}_U - \mathcal{U}_{e}^t$, as follows:
\begin{align}
    \label{eq3-4}
    \mathcal{U}_{a}^t &= \{{u}_i,\tilde{y}_{i}^t|\max(p_i^t) < \tau_{e}^t\} 
    % \notag \\ &
    \cap \{{u}_i,\tilde{y}_{i}^t|\mathrm{CG}_{i}^{t} \geq \tau_{a}^t\}
\end{align}
where $\tau_{a}^t$ is the ambiguity threshold at iteration $t$. Both $\tau_{e}^t$ and $\tau_{a}^t$ can be dynamically adjusted throughout the training process.
The choice of dynamic thresholding methods is flexible, and they can be easily integrated into our approach.
In this work, we use the Exponential Moving Average (EMA) for dynamically adjusting the thresholds.
Specifically, the EMA with a momentum $m$, which pays more attention on recent iterations, is defined as follows: 
\begin{align}
    \label{eq3-6}
    &{\tau}_{e}^t = m{\tau}_{e}^{t-1}+(1-m){\mu}_{e}^t, \notag \\
    &{\tau}_{a}^t = m{\tau}_{a}^{t-1}+(1-m){\mu}_{a}^t
\end{align} 
where ${\mu}_{e}^t$ and ${\mu}_{a}^t$, estimated over the batch $B_u$ in the current iteration $t$, are given by:
\begin{align}
    \label{eq3-5}
    &{\mu}_{e}^t = \frac{1}{B_u}\sum_{i=1}^{B_u}\max(p_i^t), \notag \\
    &{\mu}_{a}^t = \frac{1}{B_u}\sum_{i=1}^{B_u}\mathrm{CG}_{i}^{t}
\end{align}

\begin{table*}[ht]
  \renewcommand\arraystretch{1.25}
  \begin{center}
  \small
  \resizebox{\textwidth}{!}{
  \begin{tabular}{m{75pt}<{\centering}|m{30pt}<{\centering}m{30pt}<{\centering}m{30pt}<{\centering}|m{30pt}<{\centering}m{30pt}<{\centering}m{30pt}<{\centering}|m{30pt}<{\centering}m{30pt}<{\centering}m{30pt}<{\centering}|m{30pt}<{\centering}m{30pt}<{\centering}m{30pt}<{\centering}}
    \Xhline{1.5pt}
    \multirow{1}{*}{Dataset}  &\multicolumn{3}{c|}{\emph{CIFAR10}} &\multicolumn{3}{c|}{\emph{CIFAR100}} &\multicolumn{3}{c|}{\emph{SVHN}} &\multicolumn{3}{c}{\emph{STL10}}\\\cline{1-13}
    \# Labels/Class &4 &10 &25 &4 &10 &25 &4 &10 &25 &4 &10 &25\\
    
    \Xhline{1.5pt}
    PseudoLabel &76.29\tiny{$\pm$1.08} &65.18\tiny{$\pm$0.52} &48.28\tiny{$\pm$2.10} &87.15\tiny{$\pm$0.87} &76.06\tiny{$\pm$1.12} &59.09\tiny{$\pm$0.61} &75.95\tiny{$\pm$3.39} &27.49\tiny{$\pm$1.17} &16.60\tiny{$\pm$1.13} &73.81\tiny{$\pm$0.82} &68.13\tiny{$\pm$2.02} &49.21\tiny{$\pm$1.54}\\
    %\hline
    MeanTeacher &76.93\tiny{$\pm$2.29} &55.80\tiny{$\pm$4.73} &56.06\tiny{$\pm$2.03} &90.34\tiny{$\pm$0.65} &65.48\tiny{$\pm$2.37} &61.13\tiny{$\pm$0.57} &81.94\tiny{$\pm$1.33} &13.73\tiny{$\pm$0.18} &25.10\tiny{$\pm$3.17} &70.94\tiny{$\pm$0.92}
    &64.48\tiny{$\pm$0.83} &44.51\tiny{$\pm$2.02}\\
    %\hline
    MixMatch &70.67\tiny{$\pm$1.25} &53.59\tiny{$\pm$1.81} &37.28\tiny{$\pm$0.61} &79.95\tiny{$\pm$0.29} &64.31\tiny{$\pm$1.23} &49.58\tiny{$\pm$0.62} &79.63\tiny{$\pm$5.78} &24.78\tiny{$\pm$3.14} &3.71\tiny{$\pm$0.20} &68.78\tiny{$\pm$1.54}
    &55.75\tiny{$\pm$0.73} &42.34\tiny{$\pm$0.99}\\
    %\hline
    ReMixMatch &14.50\tiny{$\pm$2.58} &11.22\tiny{$\pm$1.20} &9.21\tiny{$\pm$0.55} &57.10\tiny{$\pm$0.00} &42.79\tiny{$\pm$0.53} &34.77\tiny{$\pm$0.32} &31.27\tiny{$\pm$18.79} &6.15\tiny{$\pm$0.29} &6.38\tiny{$\pm$1.09} &46.67\tiny{$\pm$4.27}
    &30.26\tiny{$\pm$5.37} &21.49\tiny{$\pm$1.75}\\
    %\hline
    FixMatch &8.33\tiny{$\pm$1.41} &7.69\tiny{$\pm$2.35} &4.97\tiny{$\pm$0.05} &53.37\tiny{$\pm$2.15} &42.75\tiny{$\pm$1.04} &34.29\tiny{$\pm$0.43} &3.56\tiny{$\pm$2.09} &\textbf{2.01\tiny{$\pm$0.05}} &\textbf{2.04\tiny{$\pm$0.02}} &40.81\tiny{$\pm$4.64}
    &16.40\tiny{$\pm$5.32} &9.81\tiny{$\pm$1.10}\\
    %\hline
    FlexMatch  &5.21\tiny{$\pm$0.36} &5.24\tiny{$\pm$0.30} &5.00\tiny{$\pm$0.05} &50.15\tiny{$\pm$1.51} &39.48\tiny{$\pm$0.70} &33.35\tiny{$\pm$0.33} &3.36\tiny{$\pm$0.37} &3.58\tiny{$\pm$0.59} &5.02\tiny{$\pm$1.20} &24.41\tiny{$\pm$7.39}
    &\underline{12.14\tiny{$\pm$0.77}} 
    &\underline{8.71\tiny{$\pm$0.17}}\\
    %\hline
    SoftMatch &5.06\tiny{$\pm$0.02} &5.08\tiny{$\pm$0.03}  
    &\underline{4.84\tiny{$\pm$0.10}} &49.64\tiny{$\pm$1.46} &38.93\tiny{$\pm$0.40} &33.05\tiny{$\pm$0.05} &2.88\tiny{$\pm$0.93} &2.29\tiny{$\pm$0.16} &2.15\tiny{$\pm$0.05} 
    &\underline{22.36\tiny{$\pm$5.37}}
    &\textbf{11.59\tiny{$\pm$2.60}} &\textbf{8.01\tiny{$\pm$0.32}}\\
    %\hline
    FreeMatch  &\underline{4.97\tiny{$\pm$0.09}} &\underline{5.01\tiny{$\pm$0.17}} &4.85\tiny{$\pm$0.10} &49.24\tiny{$\pm$2.16} &39.18\tiny{$\pm$0.95} 
    &\underline{32.79\tiny{$\pm$0.21}} &3.79\tiny{$\pm$0.03} &4.18\tiny{$\pm$1.86} &4.09\tiny{$\pm$0.66} &26.58\tiny{$\pm$3.23}
    &12.85\tiny{$\pm$0.77} &8.78\tiny{$\pm$0.88}\\
    %\hline
    FullFlex  &6.21\tiny{$\pm$1.32} &5.07\tiny{$\pm$0.24} &4.98\tiny{$\pm$0.04} 
    &\underline{48.35\tiny{$\pm$1.18}} 
    &\underline{38.86\tiny{$\pm$0.39}} &32.85\tiny{$\pm$0.10} 
    &\underline{2.84\tiny{$\pm$0.23}} &3.71\tiny{$\pm$1.51} &4.63\tiny{$\pm$1.26} &33.28\tiny{$\pm$0.85}
    &14.75\tiny{$\pm$1.99} &8.83\tiny{$\pm$0.89}\\
    \Xhline{1.5pt}
    \textbf{CGMatch} (\textbf{Ours}) &\textbf{4.87\tiny{$\pm$0.18}} &\textbf{4.92\tiny{$\pm$0.21}} &\textbf{4.80\tiny{$\pm$0.07}} &\textbf{47.55\tiny{$\pm$2.48}} &\textbf{38.68\tiny{$\pm$0.51}} &\textbf{32.51\tiny{$\pm$0.07}} &\textbf{2.39\tiny{$\pm$0.06}} &\underline{2.18\tiny{$\pm$0.06}} &\underline{2.13\tiny{$\pm$0.05}} &\textbf{22.19\tiny{$\pm$1.81}}
    &16.23\tiny{$\pm$2.39} &9.48\tiny{$\pm$0.60}\\
    \Xhline{1.5pt}
  \end{tabular}}
  \vspace{-2mm}
  \caption{Top-1 error rate (\%) on CIFAR10, CIFAR100, SVHN, and STL10 of 3 different random seeds. The best results are  highlighted in \textbf{bold} and the second-best results are \underline{underlined}.}
  \label{Results}
\end{center}
\end{table*}

\subsection{Learning From Selections}
Based on the subset selections of FDS, we apply different training objects to learn separately from easy-to-learn samples and ambiguous samples.
For the easy-to-learn set $\mathcal{U}_{e}$, we adopt the conventional cross-entropy (CE) loss: 
\begin{equation}
    \label{eq3-7}
    \mathcal{L}_{e}=\frac{1}{|\mathcal{U}_{e}|}\sum_{i=1}^{|\mathcal{U}_{e}|}H(\tilde{y}_{i},p_\theta(y|\Omega(u_i)))
\end{equation}
where $|\mathcal{U}_{e}|$ represents the number of easy-to-learn samples. For ambiguous set $\mathcal{U}_{a}$, 
we employ a more resilient regularization strategy named generalized cross-entropy (GCE) loss~\cite{gce2018} that can avoid label noise from misleading the model and is easy to optimize. The associated loss function $\mathcal{L}_{a}$ is described below:
\begin{equation}
    \label{eq3-8}
    \mathcal{L}_{a}=\frac{1}{|\mathcal{U}_{a}|}\sum_{i=1}^{|\mathcal{U}_{a}|}(\frac{1-p_{w}(\tilde{y}_{i};{u}_i)^{q}}{q}+\frac{1-p_{s}(\tilde{y}_{i};{u}_i)^{q}}{q})
\end{equation}
where $|\mathcal{U}_{a}|$ is the size of the ambiguous set; $p_{w}(\tilde{y}_{i};{u}_i)$ and $p_{s}(\tilde{y}_{i};{u}_i)$ correspond to the prediction confidence of weakly-augmented and strongly-augmented views of image ${u}_i$ given pseudo-label $\tilde{y}_{i}$, respectively. The hyper-parameter $q\in(0,1]$ is set to 0.7 as recommended in~\cite{gce2018}. The overall loss in CGMatch can be summarized to the following form: 
\begin{equation}
    \label{eq3-9}
\mathcal{L}=\mathcal{L}_{s}+\lambda_e\cdot\mathcal{L}_{e}+\lambda_a\cdot\mathcal{L}_{a}
\end{equation}
where $\lambda_e$ and $\lambda_a$ are the balancing coefficients of $\mathcal{L}_{e}$ and $\mathcal{L}_{a}$, respectively. The full algorithm of CGMatch is shown in \cref{Alg1}.
\section{Experiments}
\label{sec:exp}
\subsection{Settings}
To evaluate the performance of the proposed method, we conducted experiments on several benchmark datasets: CIFAR10/100~\cite{cifar10/100}, SVHN~\cite{svhn}, and STL10~\cite{stl} with different quantities of labeled examples. 
In our studies, we specifically focused on settings where the number of labeled samples is relatively small, i.e., 4, 10, or 25 labeled samples per class, mimicking scenarios with very limited supervision. For convenience, we use $D$-$L$ to denote the case where dataset $D$ has $L$ labels per class.
To maintain fairness in the comparisons, all methods were trained and evaluated using the unified codebase USB~\cite{usb2022} with the same backbones and hyperparameters \footnote{We re-implement the FullFlex algorithm in the USB framework.}. 
We used WRN-28-2 for CIFAR10/100 and SVHN, and WRN-VAR-37-2 for STL10. 
We adopted SGD optimizer with the initial learning rate of 0.03 and a monument of 0.9, then the learning rate was adjusted using a cosine learning rate decay schedule with the total iteration number $T$ of $2^{20}$. 
Besides, in a batch, the number of labeled data is 64, and the number of unlabeled data is 7 times larger by following previous work \cite{sohn2020fixmatch, chen2023softmatch, wang2023freematch, chen2023fullflex}. 
We ran each method with three random seeds and report the average top-1 error rate. 

In CGMatch, we first warmed-up the model on labeled data for 2048 iterations (\ie, $t_0=2048$) and retained Count-Gap values from the most recent 1000 iterations to initialize the Count-Gap queue for each unlabeled sample. 
In the subsequent training stage, 
we set $\lambda_e$ to 1 for the cross-entropy loss associated with $\mathcal{U}_{e}$;
and $\lambda_a$ for the generalized cross-entropy loss of $\mathcal{U}_{a}$ was set to increase gradually during the early training phase, when the model's predictions are still unstable, and to become more sensitive as the model improves in the mid and late stages. 
This ensures a balanced adjustment of the impact of ambiguous data on model training. 
To achieve this, we chose a non-linear increasing function with the desired properties for our experiments: $\lambda_a=\left(\frac{t-t_0}{T-t_0}\right)^2$, where $t$ is the current iteration step. 
For SVHN, as processed in FreeMatch~\cite{wang2023freematch}, 
we restricted the confidence threshold to the range of [0.9, 0.95] during the whole training process.

\begin{figure*}[ht]
    \centering
    \begin{minipage}{0.3\textwidth}
        \centering
        \includegraphics[width=\linewidth]{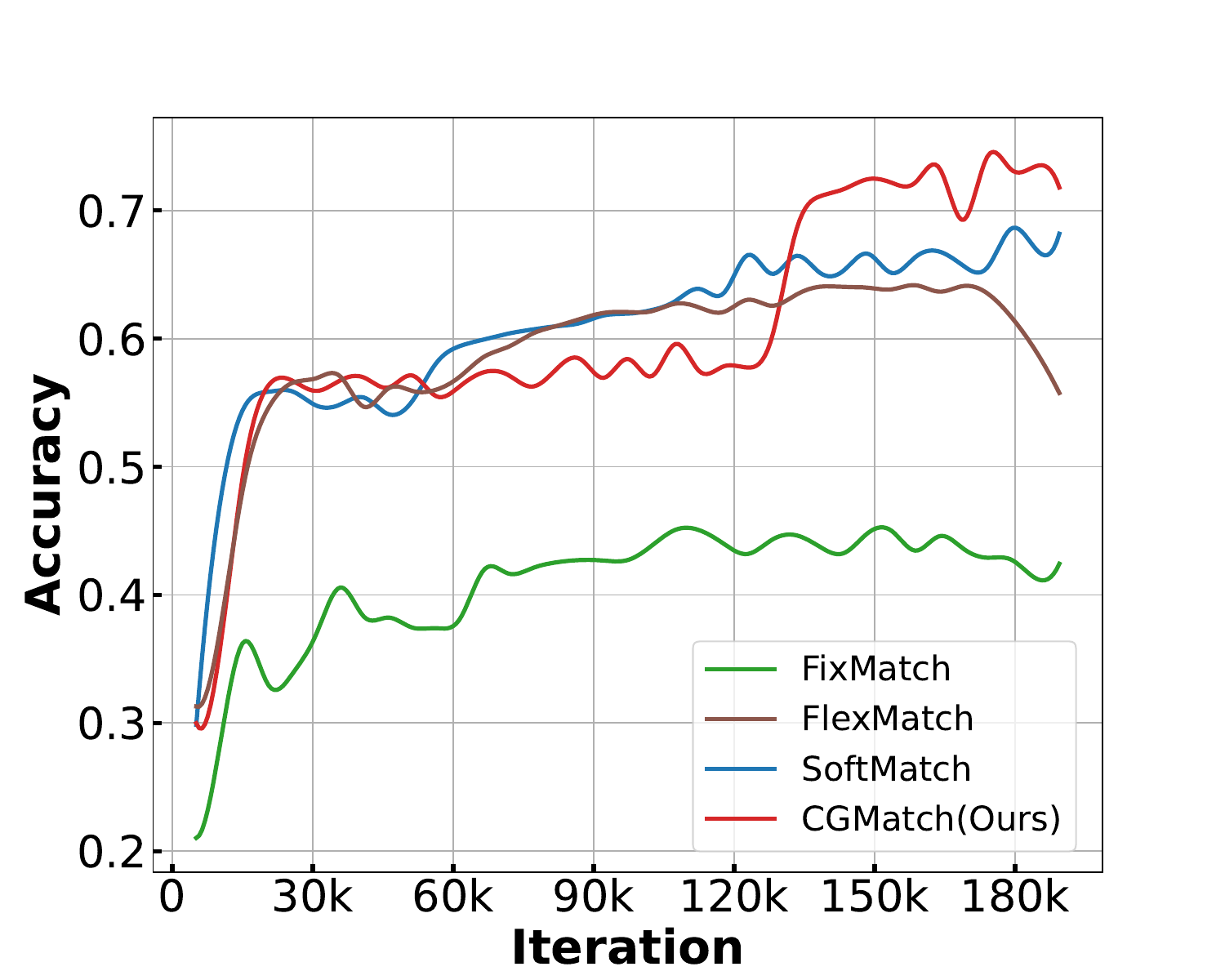}
        \subcaption{Accuracy in the early stages}
        \label{acc_early}
    \end{minipage}
    \begin{minipage}{0.3\textwidth}
        \centering
        \includegraphics[width=\linewidth]{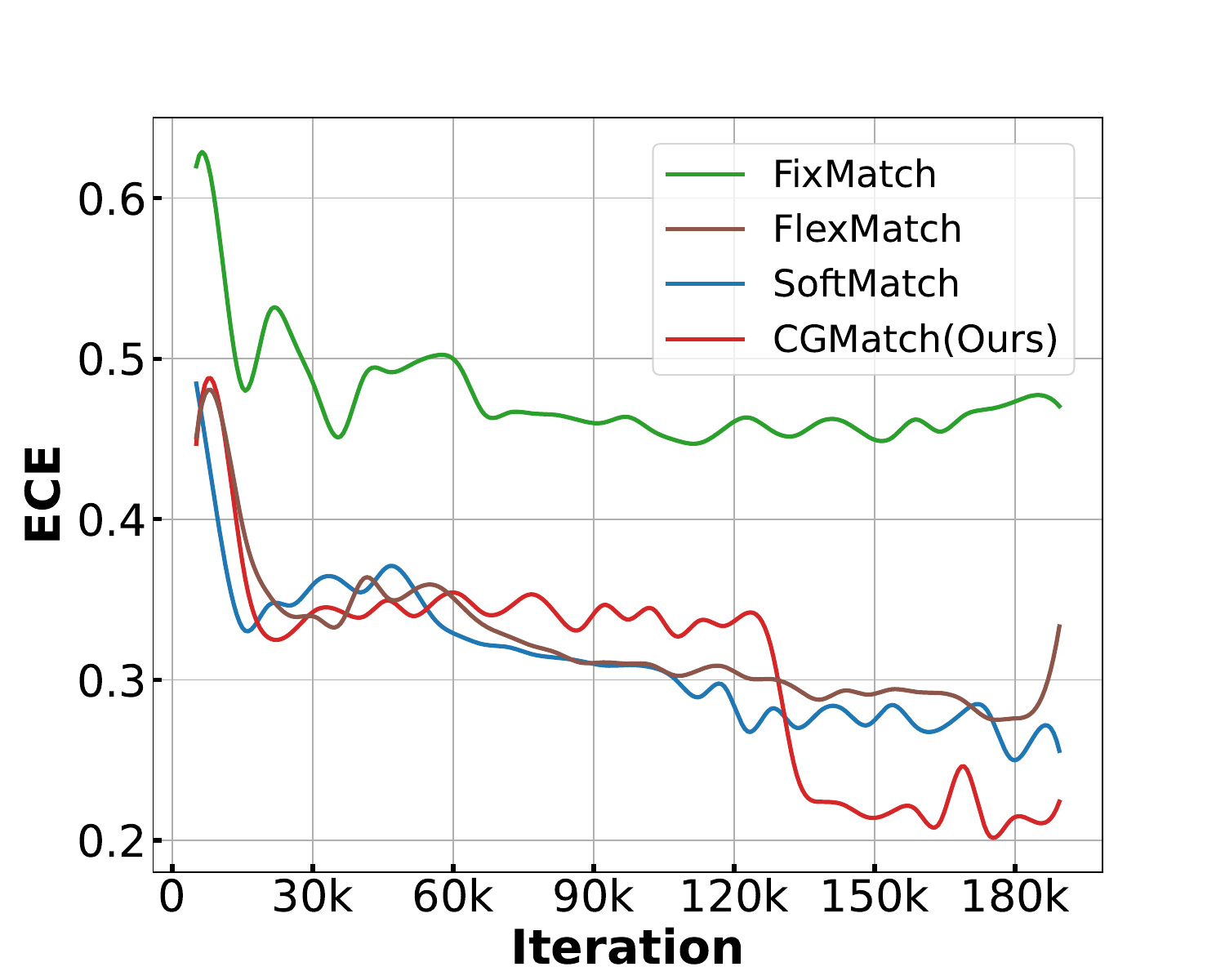}
        \subcaption{ECE in the early stages}
        \label{ece_early}
    \end{minipage}
    \begin{minipage}{0.3\textwidth}
        \centering
        \includegraphics[width=\linewidth]{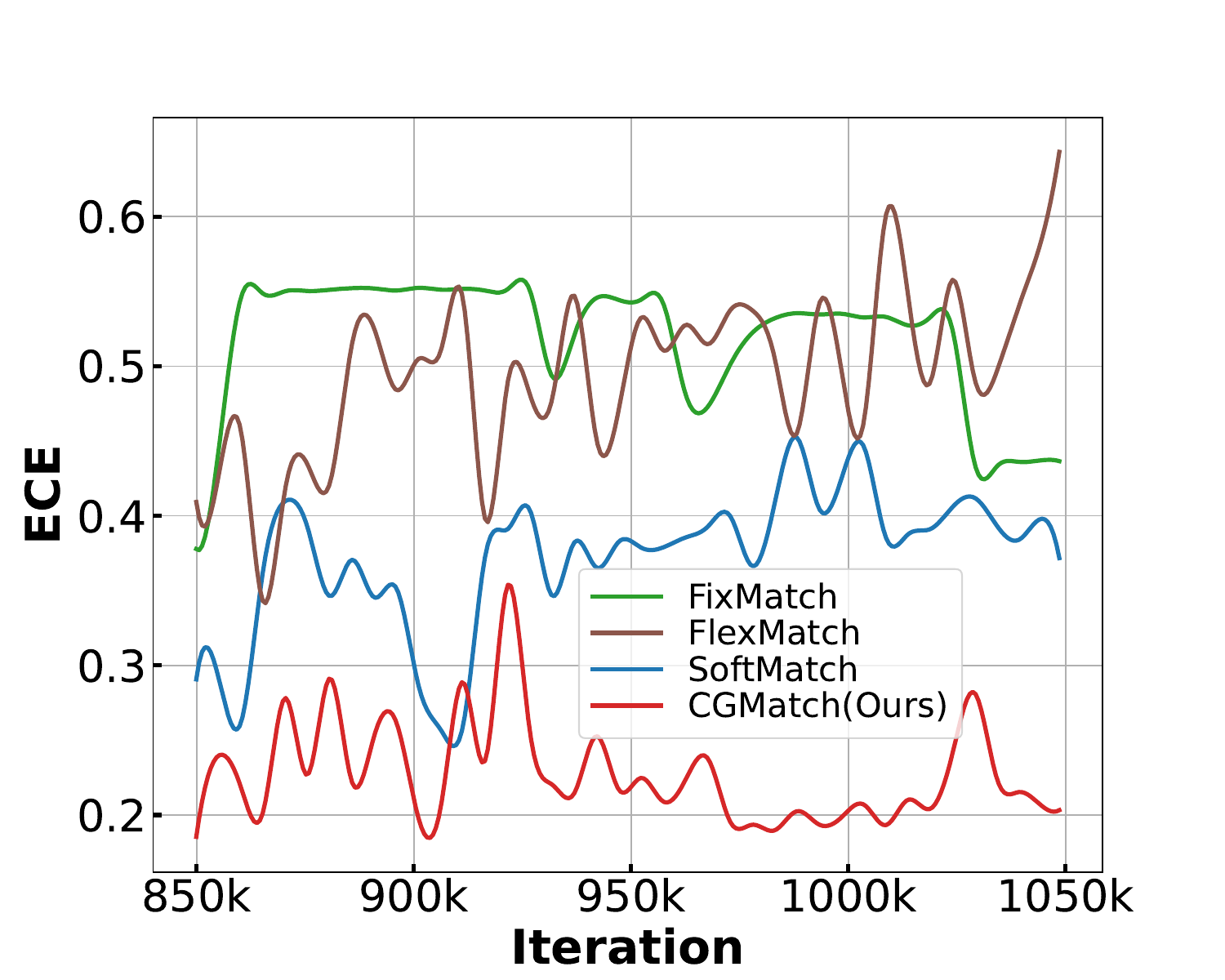}
        \subcaption{ECE in the later stages}
        \label{ece_later}
    \end{minipage}
    \vspace{-3mm}
    \caption{The learning efficiency and the model calibration error of different methods on STL10-4. (a) and (b) demonstrate the evaluation accuracy and ECE within the first 180K iterations. (c) depicts how ECE evolves over the final 200k iterations.}
    \label{acc_ece_pic}
\end{figure*}

\begin{figure}[ht]
    \centering
    \includegraphics[width=0.4\textwidth]{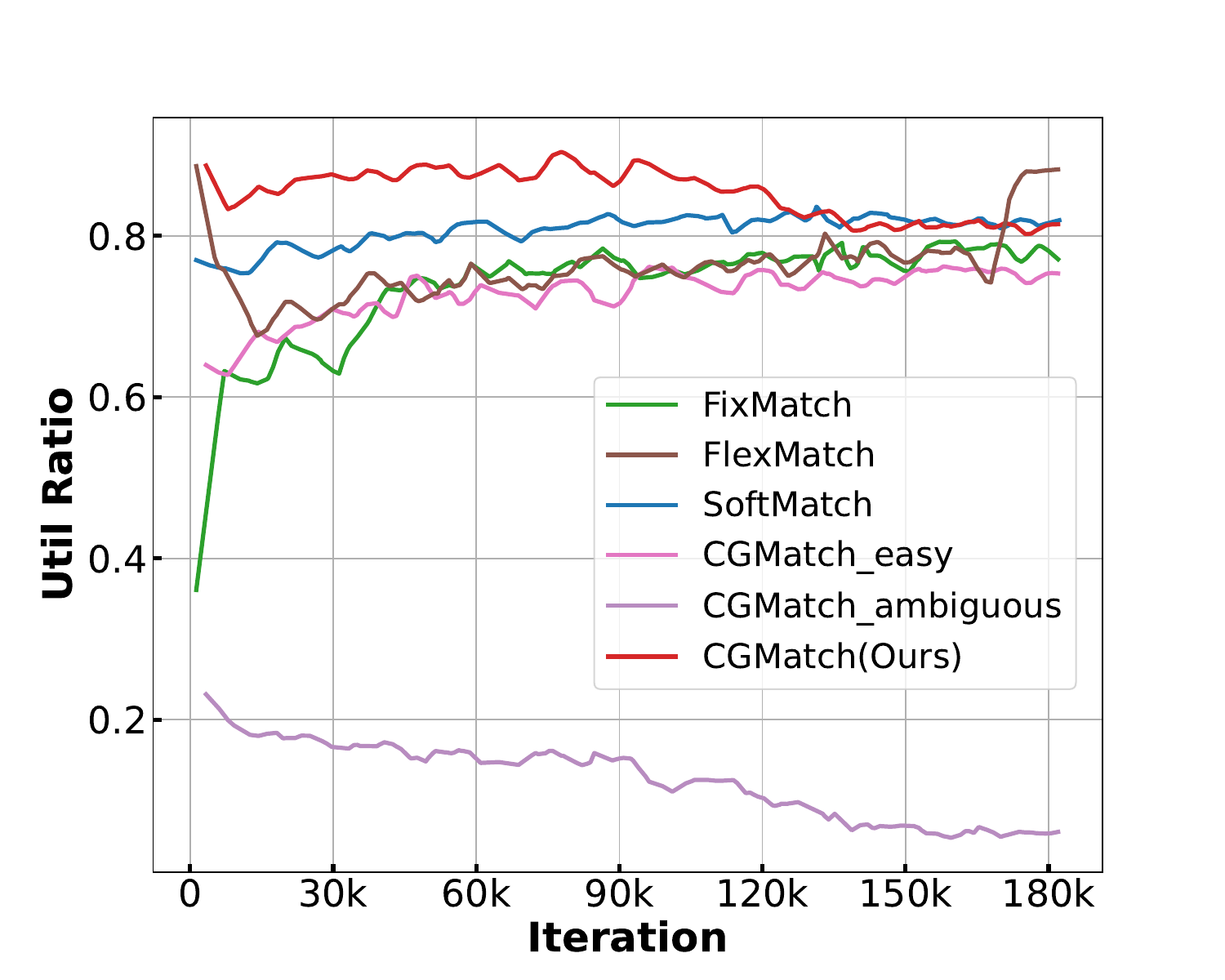}
    \vspace{-4mm}
    \caption{Comparison of the quantity of pseudo-labels on STL10-4. \textit{CGMatch}, \textit{CGMatch\_ambiguous}, and \textit{CGMatch\_easy} refer to the changing trends in the total number of unlabeled samples, the number of ambiguous samples, and the number of easy-to-learn samples involved in model training, respectively.}
    \label{util_ratio_pic}
\end{figure}
\subsection{Main Results}
\Cref{Results} presents the classification results of all compared methods. Our method performs the best on almost all cases and achieves the second-best results on SVHN-4 and SVHN-10.
We make several observations in more detail, and discuss them below. 
\begin{itemize}
    \item  CGMatch outperforms the competing methods in 8 out of 12 settings, demonstrating its effectiveness in selecting unlabeled samples that aid in model optimization and generalization, therefore further improves the classification performance. 
    \item CGMatch achieves better performance than FlexMatch in 10 out of 12 settings, particularly on settings with relatively small labeled data. For example, on more realistic dataset, CIFAR100-4, CGMatch achieves significant improvement of 2.60\%. This indicates that relying solely on dynamic confidence thresholding techniques makes it challenging to effectively utilize unlabeled data, especially when supervision is scarce, highlighting the necessity of Count-Gap. Additionally, this illustrates the capability and potential of CGMatch for deployment in real-world applications.
    \item Compared to the latest counterparts, SoftMatch and FullFlex, CGMatch delivers better classification results, except on STL10-10 and STL10-25. This implies a more effective approach should be proposed to incorporate hard-to-learn samples into training.
    \item CGMatch shows a gap compared to the best-performing method, SoftMatch, on STL10-10 and STL10-25. We argue that a possible reason is that STL10 is an open-set dataset, containing a large number of unlabeled samples from classes not present in the labeled data. After experiencing the warm-up phase, our model learns more from the known classes in the labeled data on STL10-10 and STL10-25 than on STL10-4, resulting in a larger calibration error. This makes it  challenging for Count-Gap to generalize to other unseen classes during the training phase, ultimately leading to a drop in performance.
    This can be one limitation of CGMatch, which is subject to our future work.
\end{itemize}

\subsection{Qualitative Analysis}
In this section, we provide a qualitative comparison on STL10-4 of FixMatch~\cite{sohn2020fixmatch}, FlexMatch~\cite{zhang2021flexmatch}, SoftMatch~\cite{chen2023softmatch}, and CGMatch from different aspects, and analyze why and how CGMatch works, as detailed in \cref{acc_ece_pic,util_ratio_pic}.
We employed two valuable metrics: the evaluation accuracy and the Expected Calibration Error (ECE)~\cite{guo2017ece}, and plotted their curves over iterations in \cref{acc_ece_pic} to qualify the learning status and performance of the above models.
Notably, ECE measures how well a model’s estimated “probabilities” match the true (observed) probabilities by taking a weighted average over the absolute difference between accuracy and confidence, and a lower value indicates a better-calibrated model. 
Additionally, we track the utilization ratio of unlabeled data across iterations and present their curves in \cref{util_ratio_pic}. 
The terms \textit{CGMatch}, \textit{CGMatch\_ambiguous}, and \textit{CGMatch\_easy} denote the changing trends in the total number of unlabeled samples, the number of ambiguous samples, and the number of easy-to-learn samples involved in model training, respectively. 

As illustrated in \cref{util_ratio_pic}, in contrast to these SOTA methods, CGMatch integrates not only high-confidence samples with the highest predicted probability surpassing the current confidence threshold, but also ambiguous samples into the training process, resulting in a higher utilization ratio of unlabeled data at the early stage of training. Consequently, this accelerates the training efficiency and achieves better calibration, as demonstrated in \cref{acc_early,ece_early,ece_later}, further highlighting the effectiveness of FDS scheme. For example, to reach an evaluation accuracy of 0.7, CGMatch requires only 135k iterations, significantly fewer than other methods. Then, as the model becomes more proficient and confident, ambiguous samples progressively transition into either easy-to-learn or hard-to-learn categories, leading to a reduction in the number of ambiguous samples.

\begin{figure*}[ht]
    \centering
    \begin{minipage}{0.3\textwidth}
        \centering
        \includegraphics[width=\linewidth]{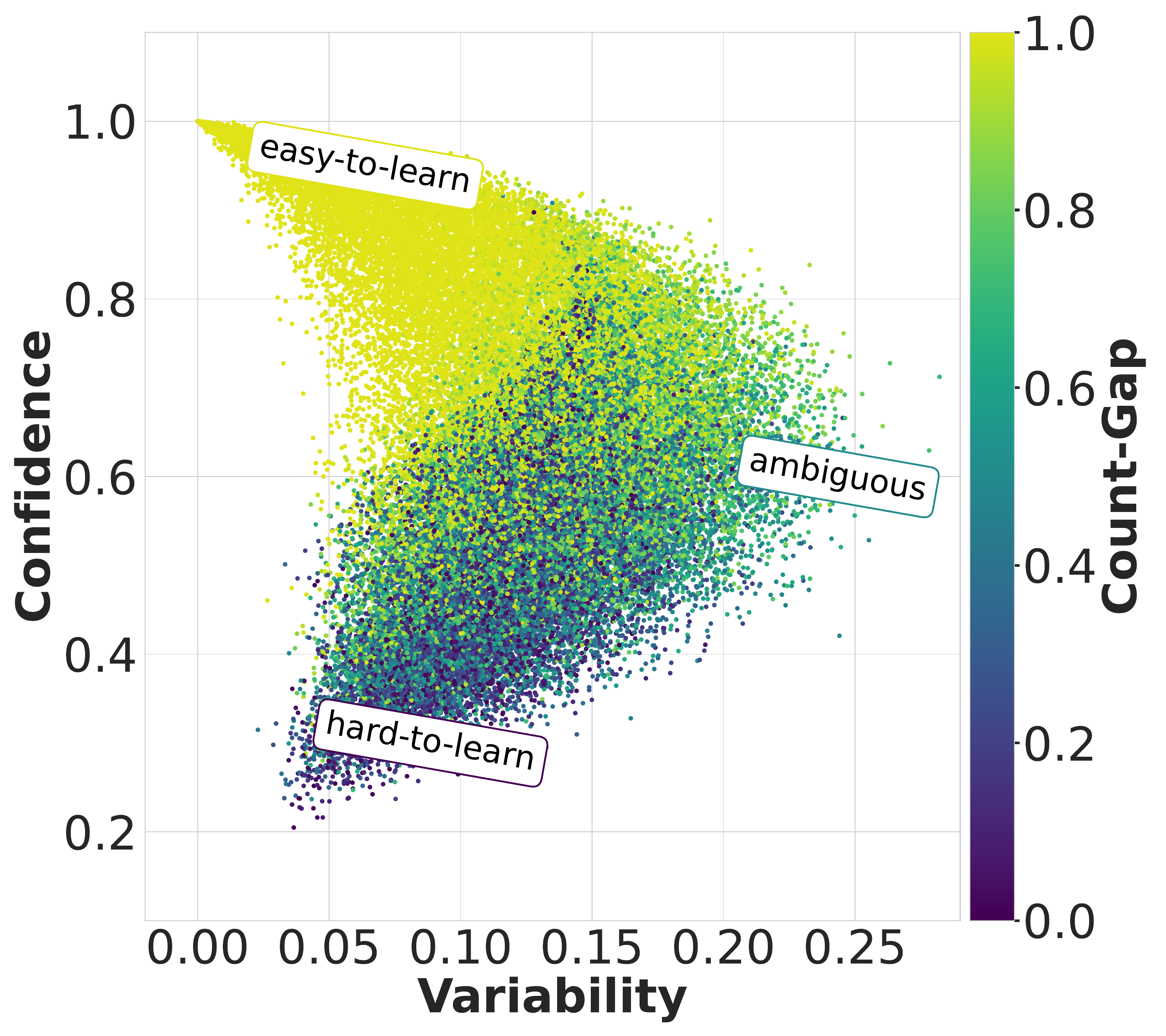}
        \subcaption{t=2048}
        \label{2048_data_map}
    \end{minipage}
    \begin{minipage}{0.3\textwidth}
        \centering
        \includegraphics[width=\linewidth]{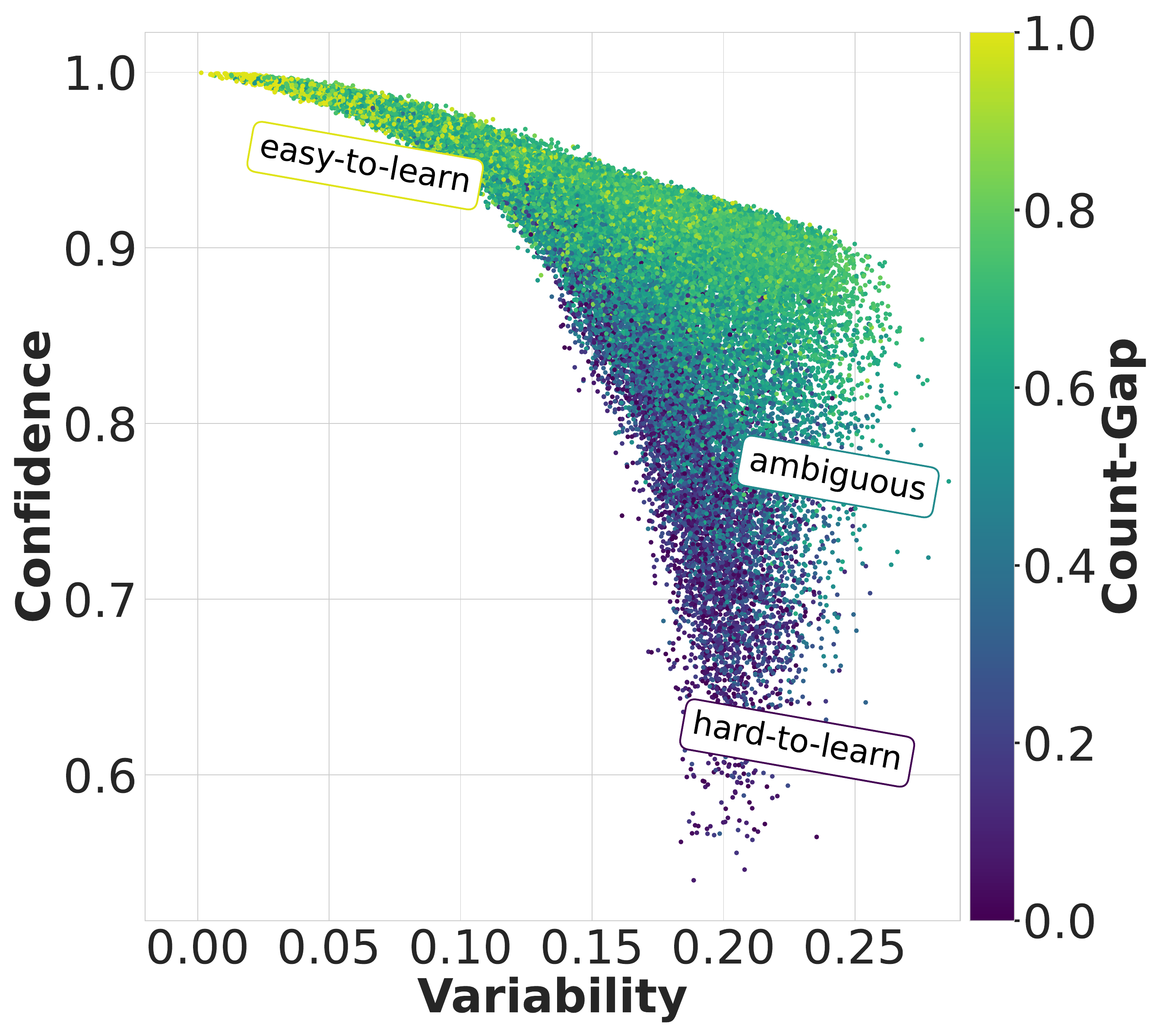}
        \subcaption{t=100k}
        \label{100k_data_map}
    \end{minipage}
    \begin{minipage}{0.3\textwidth}
        \centering
        \includegraphics[width=\linewidth]{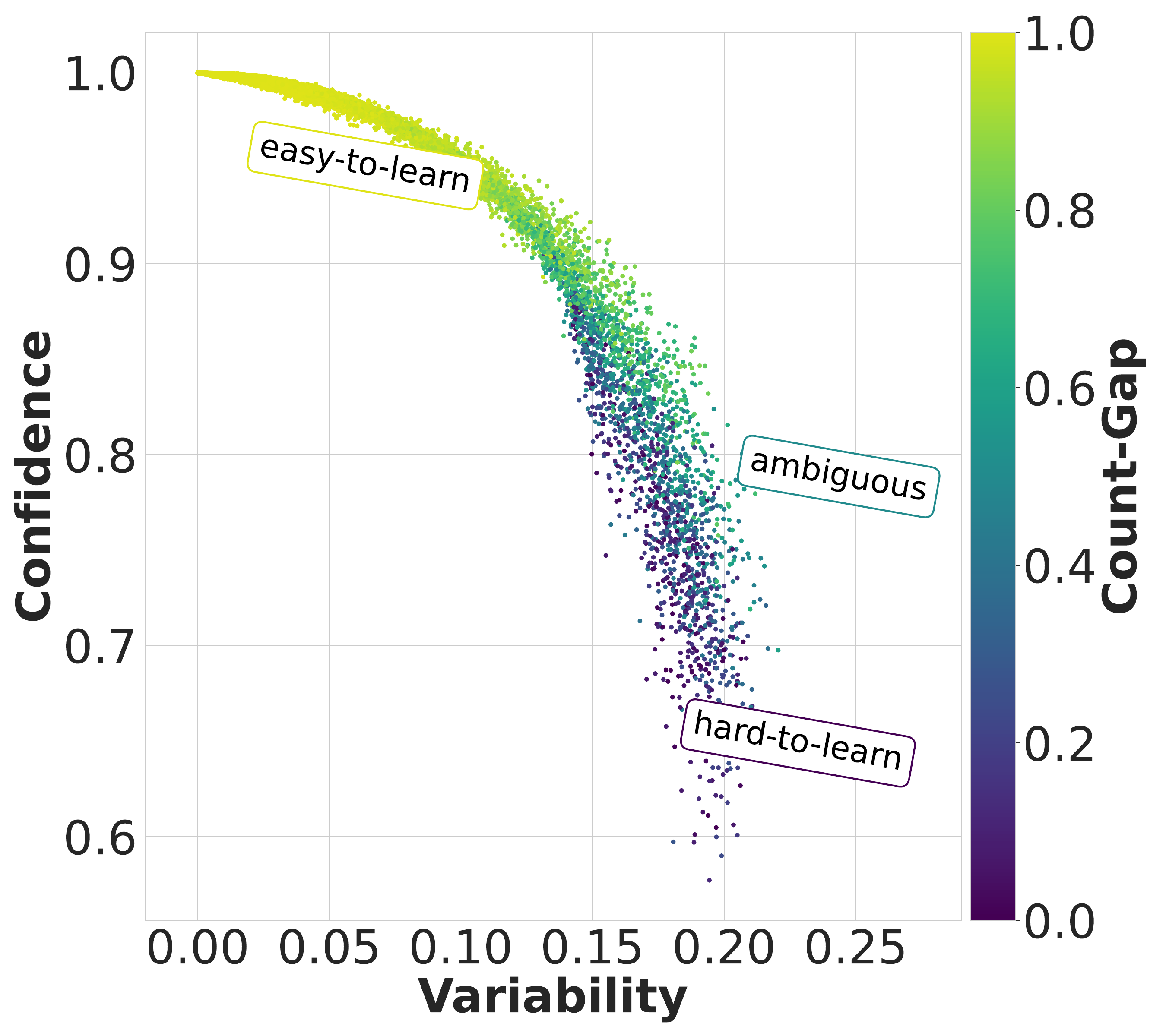}
        \subcaption{t=1024k}
        \label{1024k_data_map}
    \end{minipage}

    \vspace{-1mm}

    % 第二行单个子图
    \begin{minipage}{\textwidth}
        \centering
        \includegraphics[width=0.9\linewidth]{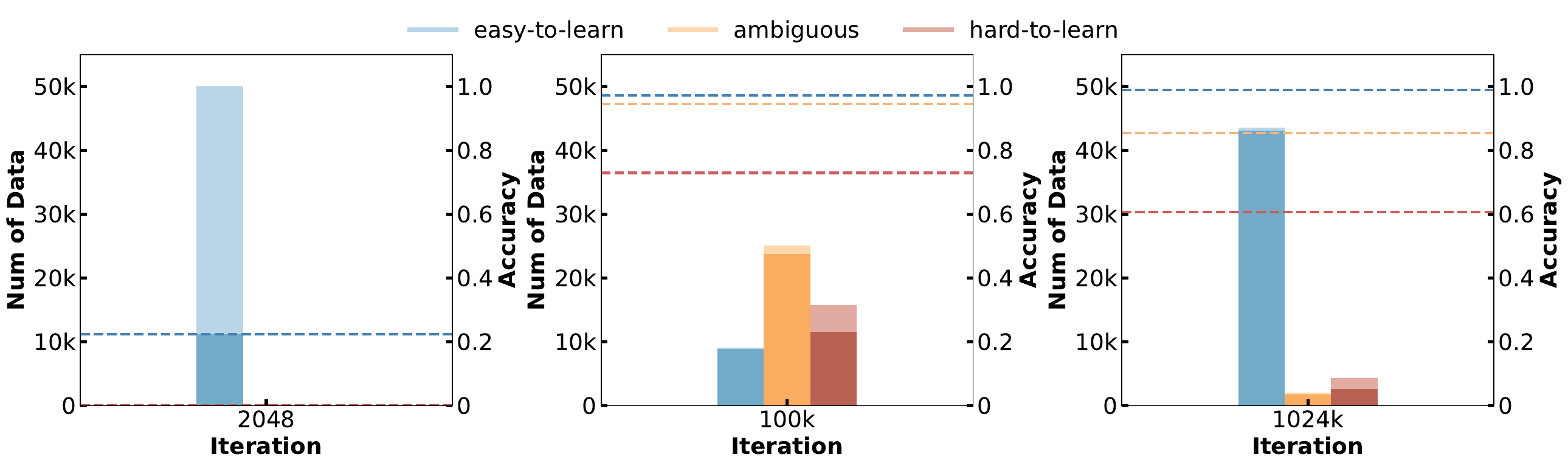}
        \subcaption{The number and prediction accuracy of easy-to-learn data, ambiguous data, and hard-to-learn data at different training iterations.}
        \label{subsets_num_acc_trends_pic}
    \end{minipage}
    \vspace{-3mm}
    \caption{Data maps and the transformation among three subsets at different training iterations of CGMatch, conducted on CIFAR10-4. At each iteration in (d), light-colored and dark-colored bars represent the total number of data and the number of correctly predicted data within each category, respectively. Additionally, the dashed lines in different colors depict the prediction accuracy for each category.}
    \label{trends_pic}
\end{figure*}

\subsection{Effectiveness Analysis of Count-Gap}
In this section, we conducted experiments to validate the effectiveness of Count-Gap. Specifically, following the data maps tool proposed in~\cite{swayamdipta2020datamaps}, we visualized the data maps regarding Count-Gap at the 2048, 100k and 1024k training iterations of the proposed CGMatch, conducted on the CIFAR10-4. 
\Cref{2048_data_map,100k_data_map,1024k_data_map} show that  as training progresses, 
the boundaries among easy-to-learn data, ambiguous data, and hard-to-learn data become  clear, 
demonstrating that Count-Gap has the ability to partition unlabeled data in the context of SSL.

Correspondingly, based on the dynamic confidence threshold and ambiguity threshold, 
we further plotted the distribution of these three types of data, along with the number of correctly predicted instances within them, across different training iterations. 
As illustrated in \cref{subsets_num_acc_trends_pic}, Count-Gap performs well in selecting high-quality unlabeled data, highlighting its effectiveness once again. 
At the beginning of the training stage (i.e. $t=2048$), 
the low confidence threshold groups all unlabeled data into an easy-to-learn set, 
and the model’s prediction accuracy is relatively low. 
As the model progressively leans form the data 
% As knowledge is gradually acquired from the data 
(i.e. $t=100k$ and $t=1024k$), Count-Gap enables increasingly precise data partitioning. This ensures that high-accuracy easy-to-learn and ambiguous samples are incorporated into the training, while hard-to-learn samples, often linked to incorrect predictions, are excluded, 
thereby reducing the negative impact of noisy labels on model performance.

\subsection{Ablation Study}

\noindent\textbf{Dynamic Thresholding Techniques.}
To verify the effect of different thresholding techniques, we consider generating the confidence threshold ${\tau}_{e}^t$ defined in our method in an adaptive or global manner, and report the mean and variance of top-1 error rate of four datasets in \cref{Ablation Study}.

\begin{itemize}
    \item \textit{Self-Adaptive Thresholding (SAT)}: Proposed in FreeMatch~\cite{wang2023freematch}, which can automatically adjust the confidence threshold for each class by leveraging the model predictions during training. 
    \item \textit{Global Thresholding (GT)}: Used in SoftMatch~\cite{chen2023softmatch}, only estimated as the exponential moving average (EMA) of the confidence from unlabeled batch at each training step.
\end{itemize}
As illustrated in \cref{Ablation Study}, \textit{GT} demonstrates superior performance across the majority of benchmark datasets when compared to \textit{SAT}. This is because accurate estimation for class-wise thresholds requires many predictions for each class, which is not available for \textit{SAT} in our settings.

\begin{table*}[ht]
  \renewcommand\arraystretch{1.25}
  \begin{center}
  \small
  \resizebox{\textwidth}{!}{
  \begin{tabular}{m{100pt}<{\centering}|m{30pt}<{\centering}m{30pt}<{\centering}m{30pt}<{\centering}|m{30pt}<{\centering}m{30pt}<{\centering}m{30pt}<{\centering}|m{30pt}<{\centering}m{30pt}<{\centering}m{30pt}<{\centering}|m{30pt}<{\centering}m{30pt}<{\centering}m{30pt}<{\centering}}
    \Xhline{1.5pt}
    \multirow{1}{*}{Dataset}  &\multicolumn{3}{c|}{\emph{CIFAR10}} &\multicolumn{3}{c|}{\emph{CIFAR100}} &\multicolumn{3}{c|}{\emph{SVHN}} &\multicolumn{3}{c}{\emph{STL10}}\\\cline{1-13}
    \# Labels/Class &4 &10 &25 &40 &10 &25 &4 &10 &25 &4 &10 &25\\
    
    \Xhline{1.5pt}
    \textit{Self-Adaptive Thresholding}  &5.19\tiny{$\pm$0.22} &4.98\tiny{$\pm$0.24} &4.83\tiny{$\pm$0.06} &48.01\tiny{$\pm$3.06} &38.83\tiny{$\pm$0.86} &33.16\tiny{$\pm$0.31} &3.59\tiny{$\pm$0.00} &4.03\tiny{$\pm$0.74}
    &5.57\tiny{$\pm$1.72} &26.75\tiny{$\pm$4.12}
    &16.38\tiny{$\pm$2.19} &10.68\tiny{$\pm$1.00}\\
    \Xhline{1.5pt}
    \textit{Global Thresholding} &\textbf{4.87\tiny{$\pm$0.18}} &\textbf{4.92\tiny{$\pm$0.21}} &\textbf{4.80\tiny{$\pm$0.07}} &\textbf{47.55\tiny{$\pm$2.48}} &\textbf{38.68\tiny{$\pm$0.51}} &\textbf{32.51\tiny{$\pm$0.07}} &\textbf{2.39\tiny{$\pm$0.06}}
    &\textbf{2.18\tiny{$\pm$0.06}}
    &\textbf{2.13\tiny{$\pm$0.05}} &\textbf{22.19\tiny{$\pm$1.81}}
    &\textbf{16.23\tiny{$\pm$2.39}}
    &\textbf{9.48\tiny{$\pm$0.60}}\\
    \Xhline{1.5pt}
  \end{tabular}}
  \vspace{-2mm}
  \caption{Top-1 error rate (\%) with different dynamic thresholding techniques on CIFAR10, CIFAR100, SVHN, and STL10. The best results are highlighted in \textbf{bold}.}
  \label{Ablation Study}
\end{center}
\end{table*}
\section{Related Work}
\label{sec:rel}

\subsection{Self-training and Pseudo-labeling}
Self-training is a commonly used technique in semi-supervised learning, centered around the iterative use of model predictions to enlarge the training dataset \cite{mcclosky2006,rosenberg2005,scudder1965,xie2020}. 
A variation of this approach, known as pseudo-labeling~\cite{lee2013}, involves converting model predictions into hard artificial labels for further training. 
Although they offer an advantage of simplicity, both of them are prone to generating noisy pseudo-labels, particularly in the early stages of training, leading to poor performance. In recent years, a variety of works mainly focus on improving pseudo-labeling by enhancing the accuracy and robustness of generated pseudo-labels \cite{sohn2020fixmatch,xie2020uda,rizve2021ups} or further taking the utilization ratio of unlabeled data into account \cite{zhang2021flexmatch,xu2021dash,wang2023freematch, chen2023fullflex, chen2023softmatch, 2023marginmatch}. 

\begin{itemize}

\item \textbf{Quality-aware Pseudo-labeling} Confidence thresholding techniques \cite{sohn2020fixmatch,xie2020uda} ensure the quality of pseudo-labels by only retaining unlabeled data with confidence levels exceeding the confidence threshold. 
Both FixMatch~\cite{sohn2020fixmatch} and UDA~\cite{xie2020uda} employ a fixed predefined threshold to select high-quality pseudo-labels. 
The difference between FixMatch and UDA is that FixMatch adopts one-hot ‘hard’ labels instead of sharpened ‘soft’ pseudo-labels with a temperature. 
Rather than solely relying on the model’s confidence at the current iteration to decide if an unlabeled example should be used for training or not, 
some works involve additional selection criteria. 
For example, UPS~\cite{rizve2021ups} combines both the confidence and uncertainty of a network prediction, with the uncertainty measured through MC-Dropout~\cite{gal2016dropout}, significantly reducing noise in the training process. 
However, the aforementioned methods inevitably face the issue of low data utilization.

\item \textbf{Quantity-aware Pseudo-labeling} Nowadays, the quantity-quality trade-off of pseudo-labeling become a trending topic in semi-supervised learning, with dynamic thresholding techniques \cite{zhang2021flexmatch,xu2021dash,berthelot2022adamatch,wang2023freematch, 2023marginmatch} or exploration strategies of the low-confidence unlabeled data \cite{chen2023fullflex, chen2023softmatch} being proposed. 
Dash~\cite{xu2021dash} proposes to gradually grow the fixed global threshold as the training progresses with an ad-hoc threshold adjusting scheme to improve the utilization of unlabeled data. 
FlexMatch~\cite{zhang2021flexmatch} considers the varying learning difficulties across different classes and defines class-wise thresholds derived from a predefined fixed global threshold; 
FreeMatch~\cite{wang2023freematch} designs a self-adaptive threshold-adjusting scheme which can reflect the model's learning status, leading to a significant performance improvement. 
MarginMatch~\cite{2023marginmatch} utilizes the model’s training dynamics on unlabeled data to improve pseudo-label data quality, 
while enabling the use of a large set of unlabelled data for learning. 
Additionally, SoftMatch~\cite{chen2023softmatch} and FullFlex~\cite{chen2023fullflex} leverage low-confidence unlabeled samples by weighting all the unlabeled samples or assigning the additional negative pseudo-label for all unlabeled data. 
Compared to these existing methods, our method achieves a better balance between quantity and quality, particularly in the challenging scenarios with only a few labeled data.

\end{itemize}

\subsection{Consistency Regularization}
Consistency regularization draws inspiration from the idea that a classifier should output the same class distribution for an unlabeled example even when it is subject to random perturbations such as data augmentation \cite{berthelot2019mixmatch,berthelot2019remixmatch,sohn2020fixmatch}, dropout~\cite{sajjadi2016dropoutcr} or adversarial perturbations~\cite{2017adversarialcr}, which serves as a valuable component in most successful SSL approaches \cite{xie2020uda,sohn2020fixmatch,chen2023fullflex,berthelot2019mixmatch,berthelot2019remixmatch}. 
Current state-of-the-art methods always (\eg, \cite{berthelot2019remixmatch,berthelot2019mixmatch,sohn2020fixmatch,chen2023softmatch}) adhere to a paradigm where training targets for unlabeled data are first generated using weakly-augmented strategies, followed by enforcing prediction consistency against the strongly-augmented versions. Similar to these approaches, our approach uses the same combination of weak and strong data augmentations.
\section{Conclusion}
\label{sec:con}
In this paper, we concentrated on the SSL settings with relatively small labels, which more accurately depict the tough yet common scenarios encountered in real-world applications. 
We have presented a novel semi-supervised model named CGMatch, designed to tackle this challenge by introducing a unique metric called Count-Gap (CG), which assesses the quality of the generated pseudo-labels from an alternative perspective. 
Based on the Count-Gap and confidence, we proposed a fine-grained dynamic selection (FDS) strategy to categorize the unlabeled data into three subsets with pronounced characteristics during the model training with the help of dynamic thresholding techniques. 
Subsequently, we adopted various regularization techniques tailored to each subset to effectively manage noise in the learning process. 
CGMatch achieves competitive performance in most scenarios. 
Moreover, it is a versatile approach applicable to most SSL frameworks and adds nearly no extra hyperparameters. 
In the future, we will aim to apply Count-Gap more effectively in open-set SSL to overcome the limitation of current method. 
Also, we will explore how to leverage hard-to-learn data to further boost the performance of SSL, rather than discarding them.

\section{Acknowledgments}
We would like to thank the anonymous reviewers for their valuable and helpful comments. This work is supported by the National Key Research and Development Program of China (No.2023YFF0905400), the National Natural Science Foundation of China (No.U2341229) and the Reform Commission Foundation of Jilin Province (No.2024C003).
\newpage
{
    \small
    \bibliographystyle{ieeenat_fullname}
    \bibliography{main}
}


\end{document}